%% file: main.tex
\documentclass[10pt,twocolumn,letterpaper]{article}
\usepackage{colortbl}
\usepackage{caption}
\usepackage[utf8]{inputenc} 
\usepackage[T1]{fontenc}    
\usepackage{booktabs}       
\usepackage{amsfonts}       
\usepackage{nicefrac}       
\usepackage{microtype}      
\usepackage{times}
\usepackage{epsfig}
\usepackage{graphicx}
\usepackage{amsthm,amsmath,amssymb}
\usepackage{mathrsfs}
\usepackage{multirow}
\usepackage{wrapfig}
\usepackage{pifont}
\usepackage[pagenumbers]{cvpr} 

\usepackage[dvipsnames]{xcolor}

\definecolor{cvprblue}{rgb}{0.21,0.49,0.74}
\usepackage[pagebackref,breaklinks,colorlinks,citecolor=cvprblue]{hyperref}
\newcommand{\zoubo}[1]{\textcolor{black}{#1}}
\newcommand{\dec}[1]{\textcolor{blue}{#1}}
\newcommand{\inc}[1]{\textcolor{red}{#1}}
\newcommand\blfootnote[1]{%
  \begingroup
  \renewcommand\thefootnote{}\footnote{#1}%
  \addtocounter{footnote}{-1}%
  \endgroup
}
\def\paperID{10861} 
\def\confName{CVPR}
\def\confYear{2024}

\title{Language-aware Visual Semantic Distillation for Video Question Answering}

\author{Bo Zou *\\
Tsinghua University\\
Beijing, China\\
{\tt\small zoub21@mails.tsinghua.edu.cn}
\and
Chao Yang *\\
Shanghai AI Laboratory\\
Shanghai, China\\
{\tt\small yangchao@pjlab.org.cn}
\and
Yu Qiao\\
Shanghai AI Laboratory\\
Shanghai, China\\
{\tt\small qiaoyu@pjlab.org.cn}
\and
Chengbin Quan\\
Tsinghua University\\
Beijing, China\\
{\tt\small quancb@tsinghua.edu.cn}
\and
Youjian Zhao \dag\\
Tsinghua University\\
Zhongguancun Laboratory\\
Beijing, China\\
{\tt\small zhaoyoujian@tsinghua.edu.cn}
}

\begin{document}
\maketitle
\blfootnote{
$^*$ Equal contribution, $^\dag$ Corresponding author \\
\indent\indent
This work was done during an internship at Shanghai AI Lab.
}

\input{sections/0_abs}
\input{sections/1_intro}
\input{sections/2_related}
\input{sections/3_method}
\input{sections/4_exp}

\section*{Acknowledge}
This work is supported in part by the Beijing Natural Science Foundation (No. L222024), and the National Natural Science Foundation of China (No. 62394322). One of the authors, Chao Yang, is supported by the Shanghai Post-doctoral Excellent Program (Grant No. 2022234).

\clearpage
{ 
    \small 
    \bibliographystyle{ieeenat_fullname} 
    \bibliography{main} 
}
\input{sections/5_appendix}

\end{document}

%% file: sections/0_abs.tex
\begin{abstract}
Significant progress in video question answering (VideoQA) have been made thanks to thriving large image-language pretraining frameworks. Although image-language models can efficiently represent both video and language branches, they typically employ goal-free vision perception and do not interact vision with language well during the answer generation, thus omitting crucial visual cues. In this paper, we are inspired by the human recognition and learning pattern and propose VideoDistill, a framework with language-aware (i.e., goal-driven) behavior in both vision perception and answer generation. VideoDistill generates answers only from question-related visual embeddings and follows a thinking-observing-answering approach that closely resembles human behavior, distinguishing it from previous research. Specifically, we develop a language-aware gating mechanism to replace the standard cross-attention, avoiding language's direct fusion into visual representations. We incorporate this mechanism into two key components of the entire framework. The first component is a differentiable sparse sampling module, which selects frames containing the necessary dynamics and semantics relevant to the questions. The second component is a vision refinement module that merges existing spatial-temporal attention layers to ensure extracting multi-grained visual semantics associated with the questions. We conduct evaluations on various challenging video question-answering benchmarks, and VideoDistill achieves state-of-the-art performance in both general and long-form VideoQA datasets. In Addition, we verify that VideoDistill can effectively alleviate the utilization of language shortcut solutions in the EgoTaskQA dataset.
\end{abstract}

%% file: sections/1_intro.tex
\section{Introduction}
\label{sec:intro}
In recent years, large-scale video-and-language pretraining has seen remarkable progress.
Most modern video-language understanding models \cite{li2019beyond,jiang2020reasoning,
li2019visualbert,fan2019heterogeneous,lei2021less,wang2022all,bain2021frozen} independently encode uni-modality and then fuse them.
Human motion perception studies \cite{woodward1998infants,baldwin2001infants,gergely2002rational} suggest that
humans perceive motion and environments as goal-driven behavior. This discrepancy leads to several issues, especially in long-form video understanding.

\begin{figure}[t]
\centering
\includegraphics[width=0.9\linewidth]{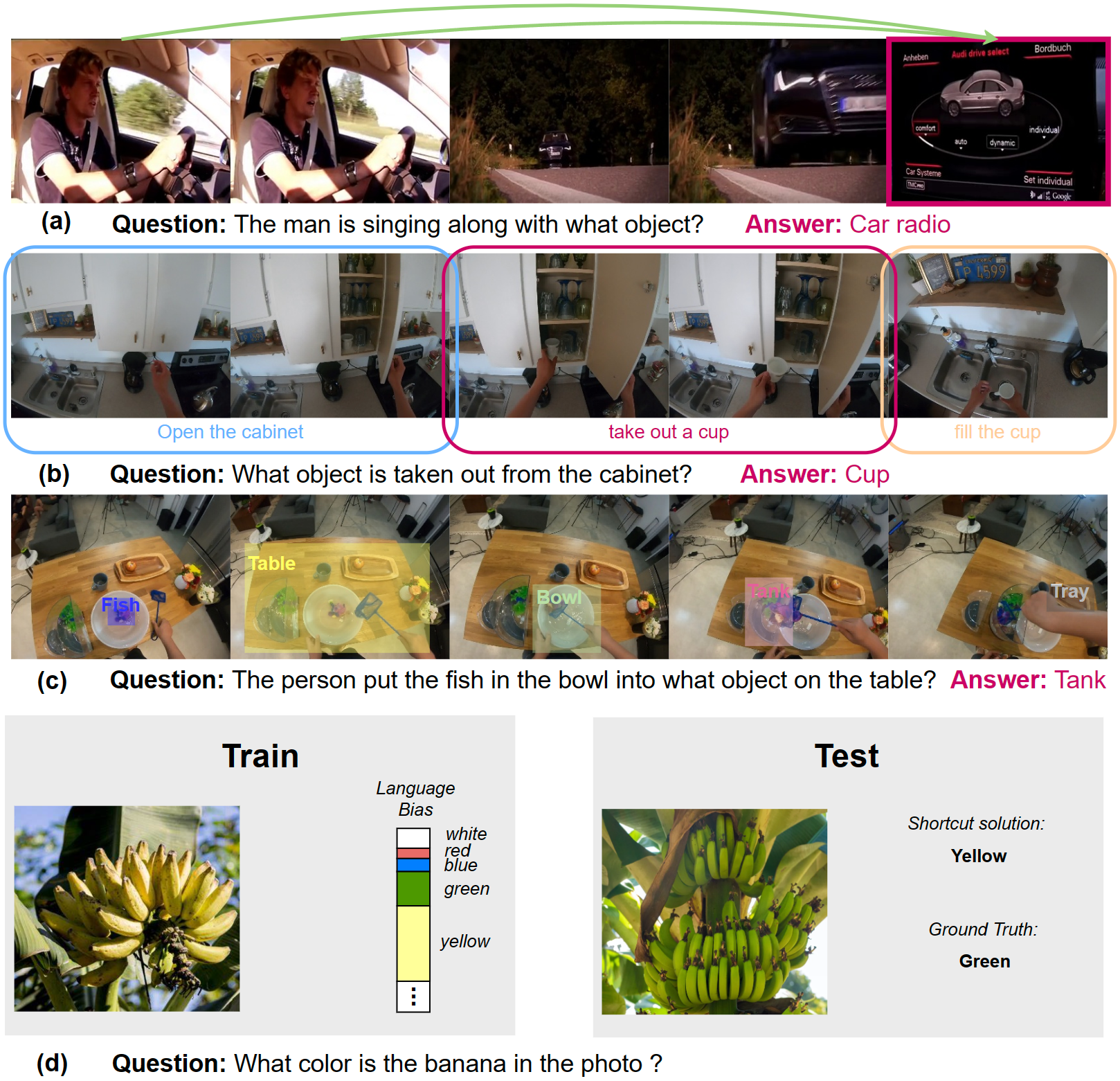}
\caption{\textbf{Challenges of goal-free VideoQA models.} They can not efficiently handle (a) Long-term dependencies, (b) Multi-events, and (c) Multi-scale semantics in the videos. They also suffer from (d) language prior phenomenon in training question-answer pairs.}
\label{figure:Challenges}
\vspace{-0.4cm}
\end{figure}

First, goal-free video representation struggles with long-term dependencies and 
multi-event reasoning.
While enabled by progress in cross-modal pretraining\cite{krishna2017visual,radford2021learning,sharma2018conceptual}
and datasets\cite{miech2019howto100m,lin2014microsoft}, 
current methods excel at question answering for images and short videos, but 
long-form videos contain many clips irrelevant or redundant to the questions, interfering with overall understanding. 
Encoding long videos also brings immense computational costs.
Second, accurate semantic reasoning usually relies on multi-scale perception from local-spatial regions to 
global temporal dynamics.
Goal-free methods for multi-scale visual modeling require custom submodels\cite{wu2022memvit, wang2021mesnet,chen2020fine} 
for each scale or extra modalities like bounding boxes and OCR features\cite{hu2020iterative, zhang2023toward} 
but large-scale pretraining makes these approaches inefficient or infeasible. 
Third, goal-free video embedding requires a multimodal fusion module to synthesize 
questions and visual embeddings to predict answers. Incorporating questions directly 
can lead to shortcut solutions \cite{niu2021counterfactual, agrawal2018don,agrawal2016analyzing,goyal2017making,chao2018cross,zhang2016yin,jia2022egotaskqa, Dancette_2021_ICCV}, which means utilizing obvious clues in questions (mainly in data distribution and the relation between keywords) that exhibit more reliability in answer prediction than complicated visual reasoning, especially in the early training phases. This phenomenon is also known as language bias, which often causes major performance gaps in out-of-distribution tests. The above challenges are visualized in Figure \ref{figure:Challenges}.

To address these issues, we propose a language-aware (goal-driven) visual semantic distillation framework called VideoDistill. Semantic distillation means the visual encoder must embed relevant frames of questions and focus on question-related multi-scale visual semantics. Closely resembles human behavior, semantic distillation functions like keeping the goal (question) in mind and conducting meaningful and precise visual reasoning. Differing from previous VideoQA frameworks \cite{seo2021look,xue2022advancing,peng2023efficient,yang2021just,li2022align,lei2021less,anonymous2024llamaexcitor}, VideoDistill can generate answers only from question-related visual embeddings (without additional textual embeddings) since the interaction of text and vision is adequate during semantic distillation. This feature also enhances the significance of visual reasoning when generating answers.

To realize semantic distillation, we first introduce Language-Aware Gate (LA-Gate), a multi-head cross-gating mechanism for cross-model interaction, which is inspired by self-gating and gated attention\cite{hu2018squeeze,li2019selective,wang2020eca, dhingra2016gated} and is integrable into any existing transformer-based models like\cite{wang2022all,bain2021frozen}. LA-Gates compute questions’ dependencies on video patch embeddings and depress or excite corresponding patches in subsequent attention layers. The proposed LA-Gate is not only the key to semantic distillation but also a new efficient modality fusion method, which is a powerful competitor of predominant Cross-Attention in VideoQA. We discuss the differences between LA-Gate and attention mechanisms in Section \ref{Language-Aware Gate} and summarize three merits of LA-Gate: $\left(1\right)$ It can alleviate the influence of language bias by avoiding the direct involvement of text embeddings. $\left(2\right)$ It can better maintain local diversity within the video embeddings. $\left(3\right)$ It enhances the interpretability of the modality fusion process.

VideoDistill has two LA-Gate-based modules. The first is a differentiable sparse sampling module, which uses pretrained image-language models like CLIP\cite{radford2021learning} to encode frames, then performs goal-driven frame sampling to remarkably reduce subsequent spatial-temporal attention overhead and naturally avoid long-term dependencies and multi-event reasoning by retaining only question-related frames. It also provides our framework with a good nature that is insensitive to the number of sampled frames (see section \ref{The Impact of Differentiable Sparse Sampling}).

The second is a vision refinement module eliminating unrelated 
visual semantics at different perceptual levels and enhancing related multi-scale semantics to 
support multi-level refinement. It encodes sparsely sampled frames into question-related global embeddings for generating answers. This module brings obvious performance boosts, especially on object-related questions (see section \ref{The Impact of Vision Refinement:}). Our contributions are summarized as follows: 

\begin{itemize}
\item We propose the Language-Aware Gate (LA-Gate) that enables interacting vision with language meanwhile not directly introducing language into visual representations. It can alleviate the influence of language bias, better maintain local diversity within the video embeddings, and is more interpretable compared with predominant Cross-Attention.
\item Based on LA-Gate, we propose a differentiable sparse sampling module to capture question-related frames and a vision refinement module to emphasize multi-scale question-related semantics. They can benefit VideoQA models on the stability under various numbers of sampled frames and the capacity to understand multi-scale objects.
\item Our model achieves new state-of-the-art performance on a wide range of downstream VideoQA tasks and text-to-video retrieval tasks.
\end{itemize}

\begin{figure*}[t]
\centering
\includegraphics[width=0.85\linewidth]{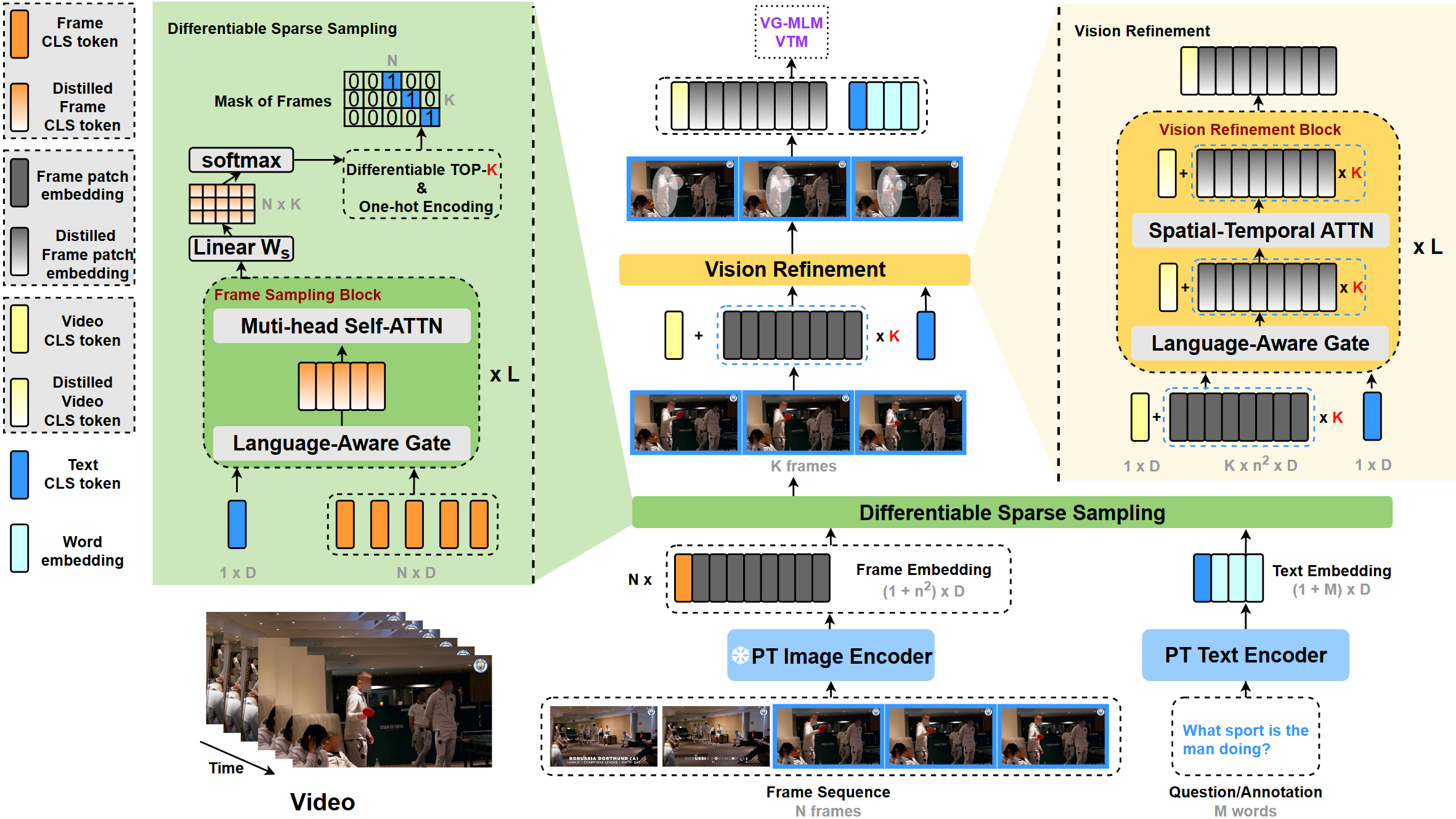}
\caption{\textbf{Overview of VideoDistill.} VideoDistill first densely samples video frames and utilizes a pre-trained image-language encoder to extract features, then sparsely samples a small number of question-related frames by a differentiable sparse sampling module. Finally, VideoDistill uses a vision refinement module to emphasize necessary multi-scale visual semantics in selected frames.}
\label{figure:2}
\vspace{-0.6cm} 
\end{figure*}

%% file: sections/2_related.tex
\vspace{-0.35cm} 
\section{Related Works}
\label{sec:related}
\subsection{Long-form Video modeling}
Long-form understanding has recently become a new research hot spot \cite{wu2019long,wu2021towards,cheng2022tallformer,xue2022advancing,gao2022mist,sun2022long,shvets2019leveraging,lei2021less,li2020hero,zellers2021merlot}. Current methods develop in two directions to overcome unique challenges not encountered by previous short-term video understanding works. 
The first direction is to encode long-term dependencies better. LF-VILA \cite{sun2022long} proposes hierarchical temporal window attention that begins with a small window learning the attention between adjacent frames, then gradually expands the window size to learn high-level representation. HERO \cite{li2020hero} and MERLOT \cite{zellers2021merlot} predict the order of shuffled frames to understand sequential characteristics.
The second direction aims to reduce the increasing computation cost of encoding longer videos. Since the common practice of randomly selecting 3 or 4 frames per video regardless of length is not suited for long videos, ClipBert \cite{lei2021less} randomly samples a sequence of segments from a video and then aggregates their predictions. 
MIST \cite{gao2022mist} improves ClipBert by selecting both video segments and frame regions to be encoded and reduces computation costs further. 
TALLFormer \cite{cheng2022tallformer} proposes short-term feature extraction and long-term memory mechanisms that avoid repetitive calculations during training. 

Similar to the second category, we reduce the workload by decreasing inputs, but our goal-driven differentiable sparse sampling works directly on frames, the minimum composition of the video, rather than \zoubo{rough} video segments. Kim et al. \cite{kim2020dense} propose to use a self-gating mechanism to realize frame sampling, but it is a goal-free sampling method. Also, unlike random selection \cite{lei2021less} or simple selector based on similarity\cite{gao2022mist}, our learnable sampler excavates the semantic relationships between visual candidates. In addition, sparsely sampled frames from our sampler can be considered a concise summary of the long video sequence. Therefore, the following inference can perform similarly to models on short-form inputs. Thus, VideoDistll does not make efforts in the first direction.

\subsection{Video-Language Fusion in VideoQA}
The answer generation of VideoQA can be formulated as the interaction of the two modalities. Some methods propose fusing language and vision in the feature extraction stage for a more interactive combination. PSAC \cite{li2019beyond} combines regular visual self-attention with visual-linguistic co-attention. HCRN \cite{le2020hierarchical} supports conditioning video features with linguistic clues as a context factor in every inference stage. PMT \cite{peng2023efficient} adopts a pyramidal video-language interaction. MCAN\cite{Yu_2019_CVPR} utilizes Guided-Attention (V2T cross-attention in this paper) to fuse language into visual feature extraction.

Unlike these techniques, our VideoDistill only keeps the question ``in mind." We do not directly fuse the question information into the video embedding, thus generating ``purer" visual embeddings. Since the answer prediction is only based on visual embeddings, VideoDistill can alleviate just assuming the answer from the question and follow the ``look and answer" criterion \cite{agrawal2018don}.

Besides VideoDistill, some existing methods can also be considered as indirect fusion. Contrastive learning methods like \cite{radford2021learning, grill2020bootstrap,chen2021exploring,wu2022rap,gorti2022x,zou2023spaceclip} pull the matched pairs closer to interact vision with language. However, they do not demonstrate the same level of performance in tasks other than retrieval \cite{kim2021vilt} due to ignoring the misalignment of vision and language. The method of fine-grained contrast to mitigate the misalignment has become a recent research hot spot.

%% file: sections/3_method.tex
\section{Methodology}
\label{sec:method}

We introduce VideoDistill, a pretraining framework that enables goal-driven VideoQA relying on question-related frames and their multi-scale semantics. Figure \ref{figure:2} gives an overview of VideoDistill. VideoDistill consists of two sub-modules: \textbf{Differentiable Sparse Sampling} and \textbf{Vision Refinement}. Both sub-modules are built upon our proposed \textbf{language-aware gate} (LA-Gate) to perform in a goal-driven manner.

\begin{figure*}[t]
\centering
\includegraphics[width=0.9\linewidth]{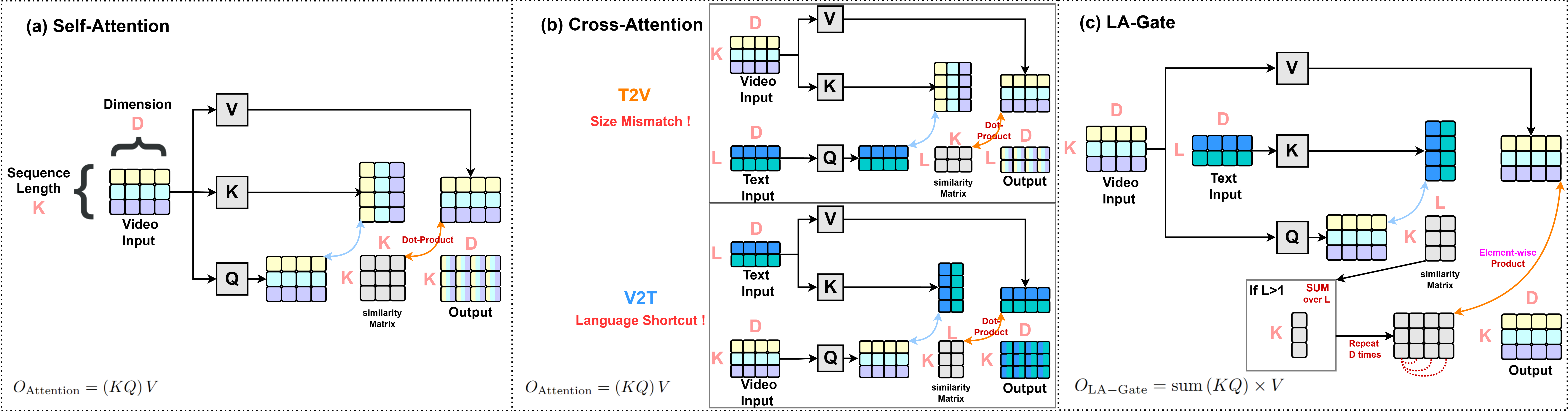}
\caption{Illustrations of Self-Attention, Cross Attention, and our LA-Gate mechanisms}
\label{figure:Gate VS.}
\vspace{-0.4cm} 
\end{figure*}

\subsection{Vision-Language Representations}
VideoDistill begins with densely sampling $N$ uniformly distributed frames and embedding them into frame representations. We adopt the pretrained image-language encoder CLIP \cite{radford2021learning} with frozen parameters in the visual branch to reduce the computational overhead. CLIP divides $N$ frames into
$N \times n\times n$ patches, and extracts patch embeddings $v_{\rm{patch}} \in \mathbb{R}^{N \times n^{2} \times D}$ and CLS tokens $v_{\rm{cls}} \in \mathbb{R}^{N \times D}$ that represent the global understanding of frames, where D is the dimension of representations. 

For the Language branch, A text is first tokenized and then fed into the pretrained text encoder of CLIP to generate word-level embeddings $t_{\rm{word}} \in \mathbb{R}^{M \times D}$ and a sentence-level token $t_{\rm{cls}} \in \mathbb{R}^{D}$, where M is the length of embeddings. Note that parameters in the text encoder are updated during pretraining and downstream finetuning to mitigate the linguistic gaps between datasets.

\subsection{Language-Aware Gate}
\label{Language-Aware Gate}
We introduce a reusable unit, termed language-aware gate (LA-Gate), which takes an arbitrary visual representation $v$ and a sentence-level language representation $t_{\rm{cls}}$ as input. LA-Gate generates $v_{\rm{distill}}$, an output sequence of language-aware visual representation with the same dimension as $v$, by exciting or depressing the components of $v$ according to their similarities with $t_{\rm{cls}}$. For a better understanding of LA-Gate, we illustrate self-attention, cross-attention, and LA-Gate in Figure \ref{figure:Gate VS.}. We first discuss why commonly used cross-attention is not suitable for our purposes, then point out the design concept and benefits of LA-Gate.

Figure \ref{figure:Gate VS.}(b) illustrates two types of cross-attention: Text-to-Video (T2V, text as $Query$) and Video-to-Text (V2T, video as $Query$). Since outputs of the attention have the same length as $Query$ and carry information sourced from $Value$, applying cross-attention in VideoQA will have two problems. Firstly, T2V is unsuitable for the modality interaction in the middle layers as it alters the shape of visual representations. Although V2T can maintain the shape, its output will carry the info directly from the text (like reconstructing text representations into the shape of visual input).

The structure of LA-Gate is shown in Figure \ref{figure:Gate VS.}(c). There are two divergences between V2T and LA-Gate. (1) When text input length $L>1$, we sum the similarity matrix to form a vector of importance for rows of the video input. $L$ functions like the number of multi-heads in the attention mechanism, and each text representation will partially dictate the importance of each row in the video input. (Note that in VideoDistill, we always have $L=1$ since the text input is $t_{\rm{cls}}$.) (2) We expand the importance vector (repeat D times) to match the shape of the video input. Finally, we apply an element-wise product rather than a dot product on the importance matrix and the Value.

The first merit of LA-Gate is the text input only controls the importance of each visual representation, and the output does not directly involve text. Thus, the answer decoder can make decisions based on "purer" visual semantics and alleviate shortcut solutions hidden in texts.

A more significant characteristic of LA-Gate in general modality fusion usage: it can better maintain local diversity within the visual input. In Figure \ref{figure:Gate VS.} (a) and (b), the output displays rows in mixed colors, representing the weighted sum of rows in the $Value$. Cross-attention will gather information from all rows in the $Value$ for each output token and present more global attributes than LA-Gate since LA-Gate does not blend all features. In Figure 4(c), we maintain the uniqueness of colors for LA-Gate in each output row because each row is produced by multiplying the corresponding row in the $Value$ and a scaler of importance. This characteristic makes LA-Gate fit more into our idea of information distillation because the language-aware distillation of a local area should be irrelevant to other regions. 

The last benefit of LA-Gate is we can enhance the interpretability of modality fusion by incorporating LA-Gate with self-attention layers. Due to the skip connection between the input and the output, commonly used V2T cross-attention brutally adds a reconstructed text (maybe linearly projected) into the visual input to fuse vision and language. The physical meaning of the structure is difficult to explain. By contrast, LA-Gate first distills the visual input by emphasizing language-related visual semantics, and then the interaction within visual representations is performed by self-attention.

The implementation of LA-Gate is visualized in Figure \ref{figure:Blocks} (we demonstrate how LA-Gate works in a vision refinement block, where $v$ consists of frame patches from $K$ selected frames and a video-level CLS token). Assume LA-Gate receives $v \in \mathbb{R}^{m \times D}$, where $m$ is the sequence length, and $t_{\rm{cls}} \in \mathbb{R}^{D}$. We first produce the key from the language representation $t_{\rm{cls}}$ and queries and values from the visual representation $v$. Then, we calculate cosine distances between each query and the key:
\begin{small}
\begin{equation}
\begin{aligned}
\mathrm{key} = \mathrm{w_{k}} \left(t_{\rm{cls}} \right), 
\mathrm{query}_{i} = \mathrm{w_{q}} \left(v_{i} \right),      
\mathrm{value}_{i} = \mathrm{w_{v}} \left(v_{i} \right)
\end{aligned}
\end{equation}

\begin{equation}
\mathrm{dist}_{i}=\mathrm{repeat} \left( \frac{\mathrm{query}_{i}\cdot \mathrm{key}}{\left\|\mathrm{query}_{i}\right\|\times\left\|\mathrm{key}\right\|} \right), \label{2}
\end{equation}
\end{small}
where $v_{i} \in \mathbb{R}^{D}$, $i \in \left[1, m\right]$ is a feature vector in $v$. $\mathrm{w_{k}}$, $\mathrm{w_{q}}$ and $\mathrm{w_{v}}$ are trainable linear projection layers, and repeat means expanding a scalar into a vector with the dimension of $D$. Since $\mathrm{dist}_{i}$ reflects the correlation between vision and language, we treat it as an importance coefficient of $\mathrm{value}_{i}$. Then, we generate distilled $v_{i}$ as formulated:
\begin{small}
\begin{equation}
v_\mathrm{distill}^{i}= \mathrm{w_{o}}\left(\mathrm{dist}_{i} \odot \mathrm{value}_{i} \right), \label{3}
\end{equation}
\end{small}
where $\odot$ denotes element-wise production and $\mathrm{w_{o}}$ is the linear output layer. In practice, we perform a standard multi-head attention setting \cite{vaswani2017attention} on $\mathrm{w_{k}}$, $\mathrm{w_{q}}$, $\mathrm{w_{v}}$ and $\mathrm{W_{o}}$, and concatenate outputs of each head to form $v_\mathrm{distill}^{i}$. Although, we only illustrate a single attention head in Figure \ref{figure:2} for simplicity. Also, we apply an input skip connection on $v_\mathrm{distill}^{i}$ for better training stability and convergence. Finally, LA-Gate outputs $v_{\rm{distill}} = \left\{ v_\mathrm{distill}^{i} | i \in \left[1, m\right]  \right\}$.

\blfootnote{
The code will be available at~\url{https://zoubo9034.github.io/VideoDistill/}
}

\subsection{Differentiable Sparse Sampling}
\begin{figure}[t]
\centering
\includegraphics[width=0.8\linewidth]{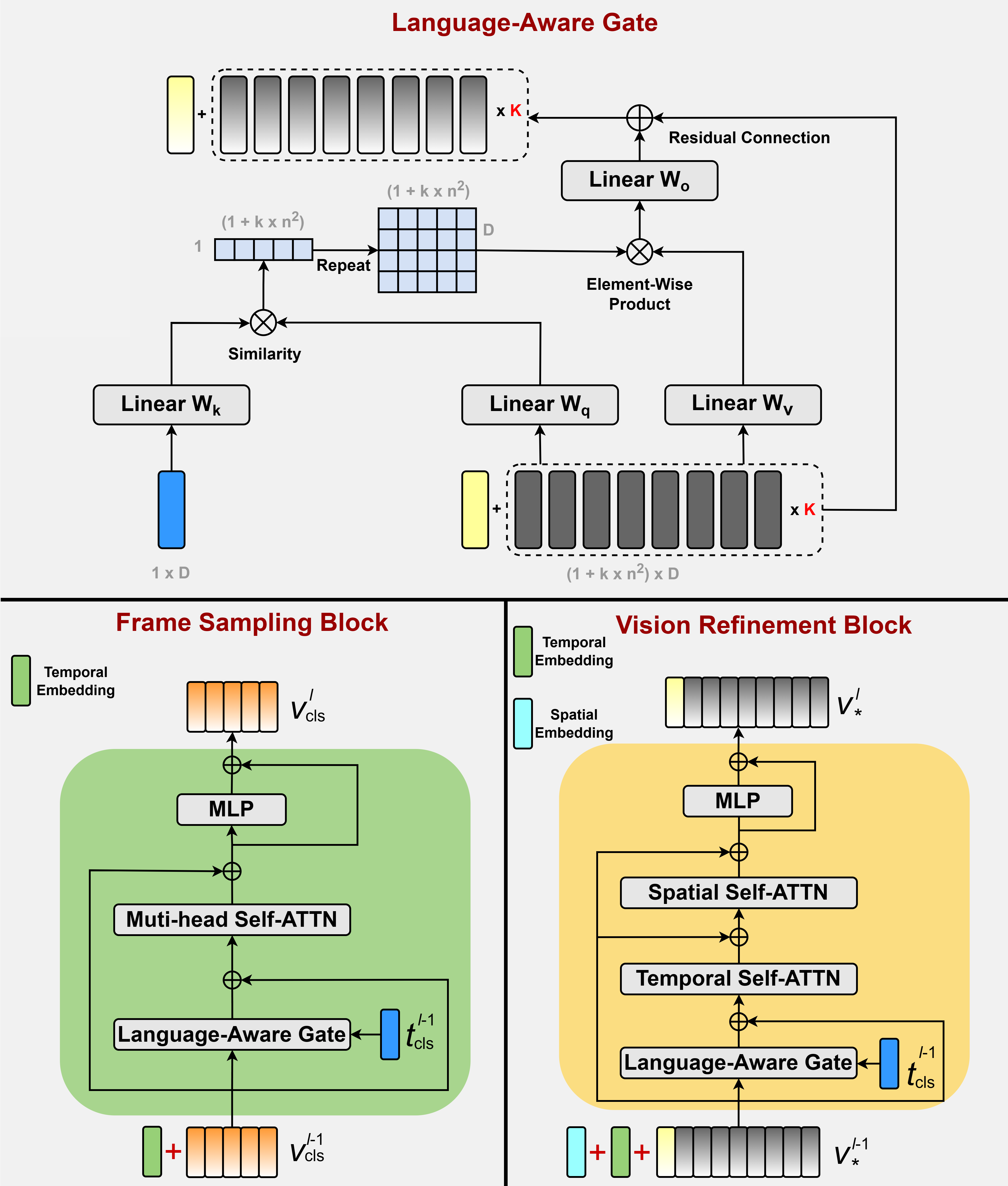}
\caption{Architectures of Language-Aware Gate, Frame Sampling Block, and Vision Refinement Block.}
\label{figure:Blocks}
\vspace{-0.4cm}
\end{figure}

Given $N$ densely sampled frames, VideoDistill further adaptively picks out  $K \left(K<N \right)$ language-related frames. To this end, we utilize stacked frame sampling blocks (FS-Blocks, $L$ layers) which take as input $v_{\rm{cls}}$ and $t_{\rm{cls}}$ to perform a top-$K$ selection. Since $v_{\rm{cls}}$ of frames are separately extracted, to indicate their temporal positions in the whole video, we add temporal embedding 
$\rm{T}_{f} \in \left\{ \phi_{\rm{T}} \left( f \right)\ | f \in \left[1, N \right]\right\}$ for each of them according to their frame index.
After adding temporal information, each FS-Block first distills $v_{\rm{cls}}$ conditioned on $t_{\rm{cls}}$ by LA-Gate to emphasize question-related frames, then performs an inter-frame interaction through standard multi-head attention.
Figure \ref{figure:Blocks} shows the architecture of FS-Blocks. We borrow the form of skip connections from the divided block in \cite{bain2021frozen}.

To realize the differentiable selection, we first project $v_{\rm{cls}}^{L}$, the output of the $L$-th FS-Block with the dimension identical to $v_{\rm{cls}}$, onto a feature space with the dimension of $K$ by a linear layer $W_{s}$, where $K$ is the number of frames to be selected. Then, we conduct Gumbel-Softmax sampling \cite{jang2016categorical} on each $x^{k} \in \mathbb{R}^{N}, k \in \left[1, K\right] $, the row of $\mathrm{W_{s}} \left( v_{\mathrm{cls}}^{L}\right)$. The procedures are formulated as follows:
\begin{small}
\begin{equation}
y_{\rm{soft}}^{k} = \mathrm{softmax} \left( \frac{x^{k} + gumbels}{\tau}  \right), \label{4}
\end{equation}
\begin{equation}
y_{\rm{hard}}^{k} = \mathrm{onehot} \left( ~\mathrm{argmax} \left( y_{\rm{soft}}^{k} \right)~ \right), \label{5}
\end{equation}
\begin{equation}
mask^{k} = y_{\rm{hard}}^{k} + y_{\rm{soft}}^{k} - \mathrm{stopgrad} \left( y_{\rm{soft}}^{k}\right), \label{6}
\end{equation}
\end{small}
\noindent
where $gumbels$ is a noise sampled from the Gumbel distribution with $\mu = 0$ and $\beta = 1$, $y_{\rm{soft}}^{k}$ is the possibility of frames to be selected as the $k$-th language-related frame and $y_{\rm{hard}}^{k}$ reflects the index of the chosen frame. Since the argmax operation has no gradient, we adopt a code-level trick in equation \ref{6} to generate $mask = \left\{ mask^{k} | k \in \left[1, K\right]  \right\}$, which can properly pass the gradient. Finally, we apply $mask$ on $v_\mathrm{patch}$ and generate $v_{\rm{patch}}^{K} \in \mathbb{R}^{K \times n^{2} \times D}$ for the following vision refinement.

\subsection{Vision Refinement}
Given the selected patch embeddings $v_{\rm{patch}}^{K}$ and the language representation $t_{\rm{cls}}$, this module generates video-level representation $v_{\rm{cls}}^{\rm{*}}$, which synthesizes multi-scale language-related visual semantics. $v_{\rm{cls}}^{\rm{*}}$ is set as a learnable token with the dimension of $D$ and is concatenated to $v_{\rm{patch}}^{K}$ as the visual input $v_{\rm{*}}$. We combine the LA-Gate and any existing spatial-temporal self-attention layer \cite{bertasius2021space, wang2022all, bain2021frozen} to form a vision refinement block as shown in Figure \ref{figure:Blocks}. When iteratively applying vision refinement blocks on $v_{\rm{*}}$, LA-Gates spontaneously distill language-related visual semantics in different perceptive fields. Since the effective perceptive field gradually expands in stacked blocks \cite{raghu2021vision}, our module can consider multi-scale objects.

Unlike predominant pyramidal models \cite{li2021proposal,yang2020temporal,peng2023efficient} for multi-scale reasoning that gathers the intermediate results of each encoding stage as final representation, we empirically adopt only the $v_{\rm{cls}}^{\rm{*}}$ from the last vision refinement block.

\subsection{Pretraining Tasks}
We adopt the two pretraining tasks to facilitate cross-modal interaction: Video-Text Matching (VTM) and Vision-Guided Masked Language Modeling (VG-MLM). 

\noindent
\textbf{Video-Text Matching}. To facilitate cross-modal interaction, we use VTM as the first pre-training task. VTM predicts whether input video-language pairs are matched. In practice, we randomly exchange the annotations of two input pairs (matched) in a mini-batch with a probability of 0.5. Then, we use a linear projection head on the top of $v_{\rm{cls}}^{\rm{*}}$ to predict two-class matching logits $y$, and formulate the VTM loss as negative log-likelihood:
\begin{small}
\begin{equation}
\mathcal{L}_{\rm{VTM}}= -\mathbb{E}_{\left(v_{\rm{cls}}^{\rm{*}}\right)} \mathrm{log} ~ p \left(y ~|~ v_{\rm{cls}}^{\rm{*}}  \right), \label{7}
\end{equation}
\end{small}
We do not adopt predominant contrastive pretraining because this paradigm requires the contrast between all possible combinations of vision and language. However, for VideoDistill, the number of possible visual embeddings of each video is proportional to the number of annotations. This leads to a quadratic growth in computational overhead (see Appendix A for more details).  Nevertheless, we still apply a contrastive constraint on matched pairs (w/o exchange) to stabilize the training:
\begin{small}
\begin{equation}
\mathcal{L}_{\rm{CL}}=-\sum_{i\in B}\log\frac{\mathbb{I}\left( \rm{matched}\right) \exp \left(s\left(v_{i},t_{i}\right)/\tau\right)
}{\sum_{j\in B} \exp\left(s\left(v_{i},t_{j}\right)/\tau\right) },\label{8}
\end{equation}
\end{small}
where B is the batch size, $v_{i}$ and $t_{i}$ are $v_{\rm{cls}}^{\rm{*}}$ and $t_{\rm{cls}}$ of $i$-th video-language pair, $\mathbb{I}$ is a indicate function of whether the video and the annotation in the $i$-th pair are matched, $s$ denotes a cosine similarity function, $\tau$ is a temperature coefficient equals 0.07 in this paper.

\noindent
\textbf{Vision-Guided Masked Language Modeling}. The commonly used MLM \cite{devlin2018bert} task aims to predict the ground truth label of the masked token according to textual context. To better map the visual and the language representations in a fine-grained manner, we propose VG-MLM, which is applied only on the matched pairs (w/o exchange) and encourages predicting the masked tokens from the visual context. In particular, we first provide VideoDistill the CLS tokens $t_{\rm{cls}}$ of unmasked annotations to sample frames and extract video representation $v_{\rm{cls}}^{\rm{*}}$. Then, we mask $I$ words in the annotations and encode the masked sentences. The outcome $w^{\rm{M}}_{i}, i \in  \left[1, I\right]$, which denotes the token of $i$-th masked word, and $v_{\rm{cls}}^{\rm{*}}$ are used to predict the $i$-th masked word. The objective of VG-MLM is formulated as follows:
\begin{small}
\begin{equation}
\mathcal{L}_{\rm{VG\text{-}MLM}}= -\mathbb{E}_{\left(t_{\rm{cls}},  v_{\rm{cls}}^{\rm{*}}\right)} \mathrm{log}  p \left(w_{i} | \mathrm{SG}\left(w^{\rm{M}}_{i}\right),  v_{\rm{cls}}^{\rm{*}}  \right), \label{9}
\end{equation}
\end{small}
\noindent
where SG means stop gradient, $w_{i}$ denote the logit of $i$-th masked word. We stop the gradient of $w^{\rm{M}}_{i}$ to enhance the importance of visual clues in language modeling. Following the heuristics of Bert \cite{devlin2018bert}, we adopt the same masking strategy. Then we adopt a two-layer MLP on top of the combination of $w^{\rm{M}}_{i}$ and $v_{\rm{cls}}^{\rm{*}}$ to generate the probability over the vocabulary, which is calculated as the negative log-likelihood loss for the masked word. The overall loss of pretraining is the combination of VTM, VG-MLM, and CL:
\begin{small}
\begin{equation}
\mathcal{L}_{\rm{total}} = \mathcal{L}_{\rm{VTM}} + \mathcal{L}_{\rm{VG\text{-}MLM}} + \mathcal{L}_{\rm{CL}}, \label{10}
\end{equation}
\end{small}



%% file: sections/4_exp.tex
\section{Experiments}
\label{sec:exp}
In this section, conduct extensive experiments utilizing the pre-trained model on downstream various types of VideoQA tasks to verify the effectiveness of our proposed VideoDistill. We also transfer VideoDistill for video-text retrieval tasks to show the generalization power of the pre-trained model in Appendix C. Our pretraining set consists of three parts: (1) 3M video-caption pairs randomly sampled from generic dataset WebVid10M \cite{bain2021frozen}. (2) 4.2M video-caption pairs randomly sampled from YouTube video dataset HD-VILA \cite{xue2022advancing}. (3) 3.8M video-caption pairs from 1-st person view dataset EgoCLIP. We report results with $N=100$, $K=16$, and $L=3$. For open-ended datasets, train a MLP classification head on the top of $v_{\rm{cls}}^{\rm{*}}$. For multiple-choice datasets, we choose the answer with the maximum logit over the VTM head. More details are in Appendix B.

\subsection{Generic VideoQA:}
We first evaluate VideoDistill on the four commonly used VideoQA datasets: \textbf{MSRVTT-QA} \cite{xu2017video}, \textbf{MSVD-QA} \cite{xu2017video}, \textbf{EgoMCQ} \cite{lin2022egocentric} and \textbf{MSRVTT-multiple-choice test} \cite{yu2018joint}. Details of each dataset and results are in the supplementary material (Appendix C).

\begin{table*}[ht]
\centering
\tabcolsep=4pt
\renewcommand{\arraystretch}{0.8}
\setlength{\abovecaptionskip}{0cm}
\caption{QA accuracies of state-of-the-art methods on AGQA v2 test set.}
\resizebox{0.8\textwidth}{!}{%
\begin{tabular}{ccccc >{\color{gray}} c >{\color{gray}}c >{\color{gray}}cc}
\toprule
Question Types       & Most Likely & PSAC  & HME   & HCRN      &AIO~\cite{wang2022all} & Temp[ATP]~\cite{buch2022revisiting} & MIST-CLIP & \bf{VideoDistill\dag}\\ \midrule
Object-relation            & 9.39        & 37.84 & 37.42 & 40.33 & 48.34  & 50.15   &51.68    & \textbf{56.32}\\
Relation-action            & 50.00       & 49.95 & 49.90 & 49.86 & 48.99  & 49.76    &    \textbf{67.18}     & \underline{55.09}   \\
Object-action              & 50.00       & 50.00 & 49.97 & 49.85 & 49.66  & 46.25     &    \textbf{68.99}      &  \underline{55.12} \\
Superlative                & 21.01       & 33.20 & 33.21 & 33.55 & 37.53  & 39.78     &    42.05         & \textbf{43.30}\\
Sequencing          & 49.78       & 49.78 & 49.77 & 49.70 & 49.61  & 48.25     &    \textbf{67.24}         & \underline{54.49} \\
Exists               & 50.00       & 49.94 & 49.96 & 50.01 & 50.81  & 51.79     &    \textbf{60.33}       &  \underline{55.74}\\
Duration comparison  & 24.27       & 45.21 & 47.03 & 43.84 & 45.36  & 49.59     &    \textbf{54.62}        & 49.08\\
Activity recognition & 5.52        & 4.14  & 5.43  & 5.52  & 18.97  & 18.96     &    \textbf{19.69}        & 10.16\\
\midrule
All                        & 10.99       & 40.18 & 39.89 & 42.11 & 48.59  & 49.79  & 54.39   &   \textbf{55.80}         \\ 
\bottomrule
\end{tabular}%
}
\label{tab:agqa}
\end{table*}

\begin{table*}[h]
\caption{Performances on EgoTaskQA \textit{normal} split. \dag denotes training from scratch.}
\centering
\resizebox{0.8\linewidth}{!}{%
\begin{tabular}{cccccccccc}
\toprule
 & Category & VisualBERT~\cite{li2019visualbert} & PSAC~\cite{li2019beyond} & HME~\cite{fan2019heterogeneous} & HGA~\cite{jiang2020reasoning} & HCRN~\cite{le2020hierarchical} & ClipBERT~\cite{lei2021less} & \bf{VideoDistill\dag} \\
\midrule
\multirow{3}{*}{\rotatebox[origin=c]{90}{Scope}}    & world       & 39.73 & 40.76 & 41.91 & 38.82 & 44.27 & 42.15 & \textbf{47.32}  \\%
                                                   & intent       & 44.51 & 46.19 & 48.92 & 42.12 & 49.77 & 40.94 & \textbf{52.53}\\%
                                                    & multi-agent & 26.29 & 30.59 & 27.98 & 23.43 & 31.36 & 27.63 & \textbf{36.95}  \\%
\midrule
\multirow{4}{*}{\rotatebox[origin=c]{90}{Type}}     & descriptive    & 41.99 & 40.63 & 41.45 & 38.04 & 43.48 & 38.45          & \textbf{47.20}  \\%
                                                    & predictive     & 30.37 & 31.98 & 35.88 & 25.57 & 36.56 & 31.50          & \textbf{40.43}  \\%
                                                    & counterfactual & 41.99 & 41.89 & 44.13 & 41.94 & 48.00 & 46.75          & \textbf{49.64}  \\%
                                                    & explanatory    & 37.42 & 37.99 & 38.85 & 35.97 & 40.60 & 42.39          & \textbf{42.53}  \\%
\midrule
\multirow{4}{*}{\rotatebox[origin=c]{90}{Semantic}} & action & 15.02 & 14.75 & 14.99 & 15.08 & 14.92 & \textbf{22.91} & \underline{16.35} \\%
                                                    & object & 23.26 & 36.53 & 36.05 & 19.09 & 45.31 & 21.80          & \textbf{54.64} \\%
                                                    & state  & 59.20 & 61.89 & 63.44 & 55.65 & 68.28 & 54.36          & \textbf{72.37} \\%
                                                    & change & 68.27 & 65.05 & 68.87 & 68.38 & 67.38 & 66.58          & \textbf{71.47} \\%
\midrule
                                                    & all    & 37.93 & 38.90 & 40.16 & 36.77 & 42.20 & 39.87 & \textbf{45.02}     \\
\bottomrule
\end{tabular}%
}
\label{tab:EgoTaskQA}
\end{table*}

\subsection{Long-Form VideoQA:} 
We evaluate our model on two recently proposed challenging datasets for the long-form VideoQA, namely EgoTaskQA\cite{jia2022egotaskqa} and AGQA\cite{grunde2021agqa}. We adopt the same finetuning setting with MSRVTT-QA and MSVD-QA. For fairness comparison, we report the results of VideoDistill w/o large-scale pertaining. Details of each dataset are in the supplementary material.

\noindent
\textbf{Results.} In the Table \ref{tab:EgoTaskQA}, our VideoDistill overwhelm previous SOTA method and bring 2.82\% overall performance gains. Remarkably, in the categories of  multi-agent and descriptive, we outperform previous work with 5.59\% and 3.72\% improvement. They show VideoDistill can better understand the storyline and temporal relations in long videos. In the category of object, we achieve 9.33\% gains, which indicates VideoDistill can capture multi-scale objects. The limited performance in the category of action is due to the strong correlations between action, object, and their change. Most methods tend to over-fit this strong language bias without thorough task understanding. Although the design of VideoDistill avoids utilizing the shortcut solutions in language. In Table \ref{tab:agqa}, The results in the gray color (AIO, Temp, and MIST-CLIP) are reported by \cite{gao2022mist}. If they follow the same evaluation metrics \cite{grunde2021agqa} as us, one of the reasons for the performance gaps in some categories is they adopt frames with higher resolution ($448 \times 448$) and more powerful encoder (CLIP-ViTB/32). Nevertheless, we still outperform current SOTA methods in Object-relation (4.64\%) and Superlative (1.25\%), which require multi-scale and multi-event reasoning. We also improve the overall performance with 1.41\% gains.

\begin{figure}[h]
\centering
\setlength{\abovecaptionskip}{0cm}
\centering
\includegraphics[width=0.85\linewidth]{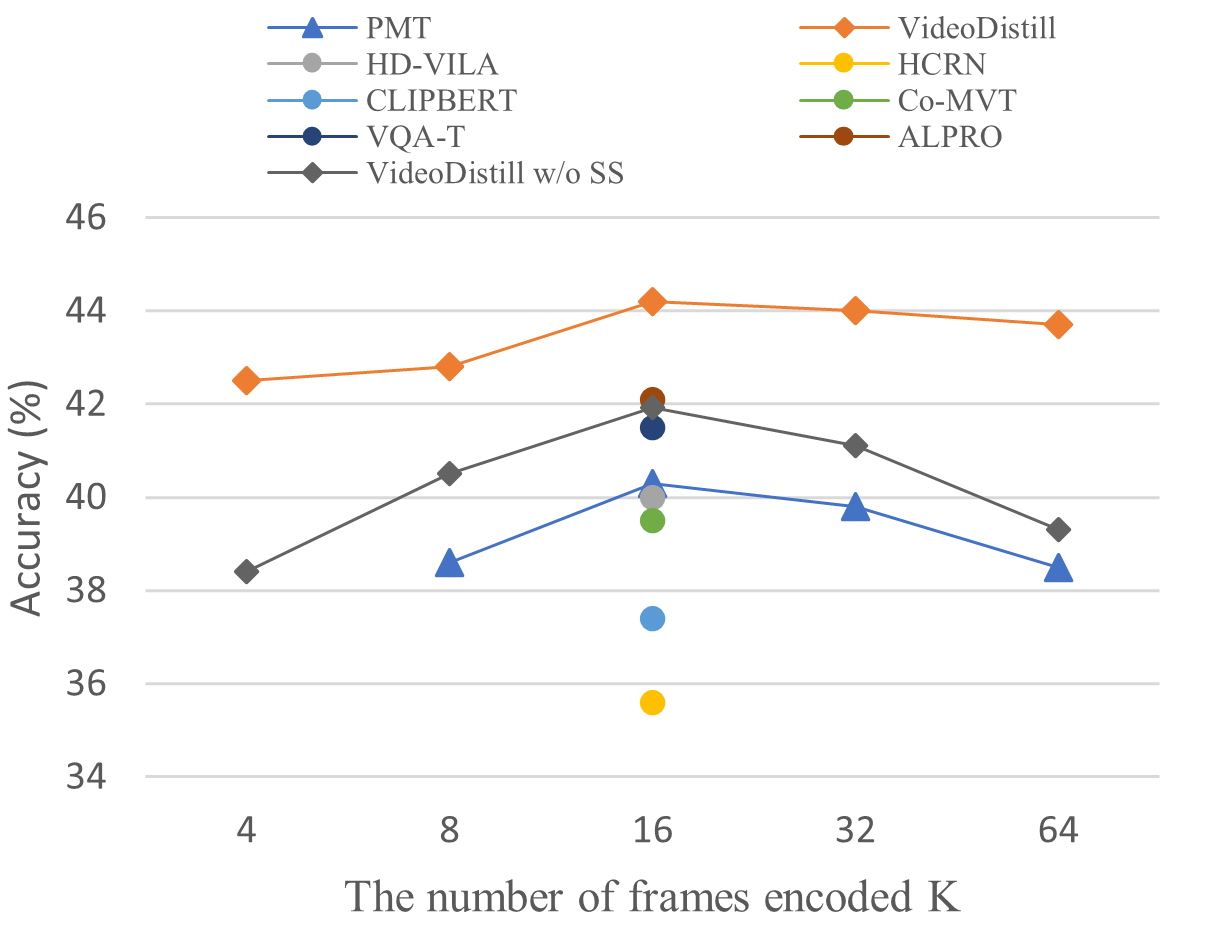}
\caption{The impact of the number of frames.}
\label{figure:Number of frames}
\vspace{-8pt}
\end{figure}

\subsection{The Impact of Differentiable Sparse Sampling}
\label{The Impact of Differentiable Sparse Sampling}
\textbf{Insensitive to the number of frames.} We evaluate the performances on MSRVTT-QA with variable input frames. In Figure \ref{figure:Number of frames}, we compare the performance change of VideoDistill (w/ \& w/o sparse sampling module) with PMT \cite{peng2023efficient} (other methods only report results on 16 frames, which are the best results), and there are two findings. First, the differentiable sparse sampling module makes our 4-frame variation on par with other methods with many more input frames. It is intuitive because other competitors require more frames to improve the change to capture necessary information for reasoning (a disadvantage of goal-free perception). This property can dramatically reduce the computational overhead required to achieve the same performance as previous methods. Second, we notice that too many frames will damage the answering accuracy since the emergence of unrelated information disturbs the reasoning process. However, our sparse sampling module can mitigate the dropping with increasing frames and make the model more robust.

\noindent
\textbf{Qualitative Results.}
We visualize the results of our sparse sampling module on EgoTaskQA in Appendix E. 

\begin{table*}[h]
\tiny
\tabcolsep=12pt
\centering
\setlength{\abovecaptionskip}{0cm}
\renewcommand{\arraystretch}{0.7}
\captionof{table}{Language-only QA results on the EgoTaskQA \textit{normal} split. \textbf{The more performance drops, the better.}}
\label{tab:without_vision}
\resizebox{0.8\textwidth}{!}{
\begin{tabular}{cccccccc}
\toprule
 \multirow{2}{*}{Category} & \multicolumn{2}{c}{BERT} & \multicolumn{2}{c}{HCRN (w/o vision)} & \multicolumn{2}{c}{\bf{VideoDistill} (w/o vision)}\\
 \cmidrule(lr){2-3}\cmidrule(lr){4-5} \cmidrule(lr){6-7}
 & Acc. & Change & Acc.& Change & Acc.& Change\\
 \midrule
world &  36.28 & \dec{-8.7\%} & 30.17 & \dec{-31.9\%} & 18.78 & \dec{-60.3\%} \\
intent & 35.02 & \dec{-21.3\%} & 35.54 & \dec{-28.6\%} & 22.57 & \dec{-57.0\%}\\
multi-agent & 20.58 & \dec{-21.7\%} & 19.9 & \dec{-36.5\%} & 11.50 & \dec{-68.7\%}\\
\midrule
descriptive & 34.55 & \dec{-17.7\%} & 29.97 & \dec{-31.1\%} & 21.19 & \dec{-55.1\%}\\
predictive & 24.75 & \dec{-18.5\%} & 18.32 & \dec{-49.9\%} & 7.42 & \dec{-81.6\%}\\
counterfactual & 41.3 & \dec{-1.6\%} & 41.1 & \dec{-14.4\%} & 39.70 & \dec{-20.0\%}\\
explanatory & 31.78 & \dec{-15.1\%} & 32.41 & \dec{-20.2\%} & 13.69 & \dec{-67.8\%}\\
\midrule
action & 15.72 & \inc{+4.6\%}  & 17.31 & \inc{+16.0\%} & 11.81 & \dec{-27.8\%}\\
object & 7.43 & \dec{-68\%} & 8.85 & \dec{-80.5\%} & 0 & \dec{-100.0\%}\\
state & 45.03 & \dec{-23.9\%} & 35.51 & \dec{-48.0\%} & 20.21 & \dec{-72.1\%}\\
change & 69.87 & \inc{+2.3\%} & 70.47 & \inc{+4.6\%} & 55.38 & \dec{-22.5\%}\\
\bottomrule
\end{tabular}
}
\setlength{\abovecaptionskip}{0.1cm}
\caption{Analysis of the effectiveness of each module.}
\label{tab:effectiveness}
\centering
\resizebox{0.8\textwidth}{!}{
\begin{tabular}{l c c c c c c}
\toprule
\multicolumn{1}{c}{\multirow{2}{*}{\bf {Method}}} & \multicolumn{4}{c}{\bf {Module}}   & \multicolumn{1}{c}{\multirow{2}{*}{\bf {EgoTaskQA}}}	 &  \multicolumn{1}{c}{\multirow{2}{*}{\bf {MSRVTT-QA}}}     \\
\cmidrule(lr){2-5}
\multicolumn{1}{c}{} & \multicolumn{1}{c}{SS} & \multicolumn{1}{c}{VR} & \multicolumn{1}{c}{Cross-ATT} & \multicolumn{1}{c}{LA-Gate} &\multicolumn{1}{c}{} &\multicolumn{1}{c}{}\\
\midrule
(a) Baseline & \ding{55} & \ding{55}  & -          & -          & 29.10 & 23.92 \\
(b) Uniform Sampling & \ding{55} & \checkmark & \ding{55}  &\checkmark  & 41.70 & 40.92 \\
(c) Soft Sampling    & \ding{55} & \checkmark & \ding{55}  &\checkmark  & 38.95 & 38.16 \\
(d) w/o VR           & \checkmark & \ding{55} & \ding{55}  &\checkmark  & 33.44 & 26.51 \\
(e) Cross-Attention  & \checkmark &\checkmark & \checkmark & \ding{55}  & 40.94 & 41.17 \\
(f) VideoDistill         & \checkmark &\checkmark & \ding{55}  &\checkmark  &\textbf{45.02} & \textbf{44.20} \\
\bottomrule
\end{tabular}%
}
\end{table*}

\subsection{The Impact of LA-Gate:}
\label{The Impact of LA-Gate}
\textbf{Prevent the use of language shortcuts.} We come up with two methods to quantitatively analyze this characteristic. First, we test the language-only QA performances on EgoTaskQA normal split in Table \ref{tab:without_vision}, which means replacing the visual inputs with a static video (We also report the performances under a Gaussian noise input in Appendix E). The more declines, the better. When compared with the language-based model Bert \cite{devlin2018bert}, and the previous SOTA method HCRN \cite{le2020hierarchical}, the decline of our VideoDistill is the most significant (especially in the object category). It demonstrates that VideoDistill relies more on vision and can avoid using language bias. Second, we test VideoDistill on EgoTaskQA indirect split, which is motivated by the fact \cite{jia2022egotaskqa} that during task execution, actions, objects, and their changes are often strongly correlated. It leaves the chance for the model to perform well by simply over-fitting these strong correlations (language bias) without thorough task understanding. The indirect references can avoid these correlations. In Appendix E, we show that our VideoDistill has the least absolute performance change. It indicates that VideoDistill barely utilizes language bias in questions.

\noindent
\textbf{A more effective fusion method than Cross-Attention.} In Table \ref{tab:effectiveness} (e), we replace LA-Gates in our frameworks with Cross-Attention. When comparing (e) with (f), we can find that LA-Gate brings significant performance boosts on both test sets (4.08\% and and 3.03\%)

\subsection{The Impact of Vision Refinement:}
\label{The Impact of Vision Refinement:}
\textbf{Performance boost in multi-scale object-related questions.} 
The stacked video distillation blocks containing LA-Gate can obtain multi-scale question-related semantics. We observe huge improvements in Table \ref{tab:EgoTaskQA} object category (9.33\%) and Table \ref{tab:agqa} object-relation category (4.64\%).

\subsection{Ablation Study}
We conduct ablation studies to verify the effectiveness of components and the rationality of parameter selection.



\noindent
\textbf{Effectiveness of each component.} We design four variants and report their performances in Table \ref{tab:effectiveness}. (a) we remove the two modules and apply evenly sampled 16 frames as input and a 3-layer transformer to fuse the vision and language. (b) we remove differentiable sparse sampling and apply evenly sampled 16 frames as input. (c) we replace the hard sampling in SS with soft sampling, which means using the weighted sum of dense frames rather than top-k selection. (d) we remove vision refinement and apply a 3-layer transformer to fuse the vision and language. (e) we replace LA-Gates in both modules with cross-attention layers.

\noindent
\textbf{More Ablations} We validate the influence of pretraining tasks, the best number of densely sampled frames, and the reasonable number of stacked layers $L$ in Appendix E.

\subsection{Conclusion}
In this paper, we study language-aware VideoQA to overcome the difficulties of long-term dependencies, multi-events, multi-scale semantics, and shortcut solutions in video understanding. In particular, we introduce a differentiable sparse sampling module that naturally avoids complicated long-term dependencies and multi-event reasoning since it only retains question-related frames and a vision refinement module that eliminates unrelated visual semantics in different perception fields and enhances multi-scale related semantics to support a multi-level refinement process. Both modules are based on our LA-Gate, which realizes cross-modal interaction meanwhile avoiding language's direct fusion into visual representations. LA-Gate is a powerful competitor of predominant Cross-Attention in VideoQA and fits our idea of semantic distillation. Therefore, we can realize high-performance VideoQA by just relying on question-related visual representations and reducing language shortcuts. Experiments on 8 VideoQA and text-to-video retrieval tasks demonstrate the effectiveness of our model.

%% file: sections/5_appendix.tex
\appendix
\tableofcontents

\section{Contrastive pretraining for language-aware models}
We explain the surge of computational overhead for language-aware models under contrastive pretraining. Since goal-free perception independently encodes vision and language, for B video-language pairs, we only need to encode B video clips and then compute a similarity matrix with a shape of $B\times B$. However, our experiments confirm VideoDistill will fast degenerate if we just contrast between matched pairs (calculate a single representation for each video based on its matched annotation and compute a $B\times B$ similarity matrix as we do in Equation 7). The reason for this phenomenon is the video encoder simultaneously takes matched video-language pairs as input. It can simply meet the requirements of the contrastive objectives if its output is always identical with language inputs, whatever video is received. To avoid degeneration, the comparison can not be limited to matched pairs. We should encode videos with all possible annotations in the mini-batch (compute $B^{2}$ video representations and $B \times B^{2}$ similarity matrix). Also, we should constrain each video representation based on an unmatched annotation to be unfamiliar with the videos' matched annotations.

Nevertheless, the full contrastive learning for language-aware models leads to a quadratic growth in computational overhead. This demand is beyond the reach of our current resources. We will further study this full contrastive learning in future work.

\begin{table}[ht]
\tiny
\setlength{\belowcaptionskip}{-0.05cm}
\center
\caption{Comparison with SOTA methods on MSRVTT-QA.}
\renewcommand\arraystretch{0.8}
\resizebox{1\linewidth}{!}{
\begin{tabular}{l c c c}
\toprule
\bf {Method}         				            &\bf {Pretraining data}	        & \bf {Pairs}           & \bf {Acc}  \\
\midrule
ST-VQA\cite{jang2017tgif}                       & - 		                    & -                     & 30.9     \\
Co-Memory \cite{gao2018motion}                  & -                             & - 		            & 32.0  \\
AMU \cite{xu2017video}          		        & -                             & - 		            & 32.5  \\
HME \cite{fan2019heterogeneous}                 & -                             & - 		            & 33.0  \\
SSML \cite{amrani2021noise}                     & HowTo100M \cite{miech2019howto100m}  & 136M           & 35.1 \\
HCRN \cite{le2020hierarchical}                  & -                             & - 		            & 35.6  \\
ClipBert \cite{lei2021less}                     & COCO \cite{chen2015microsoft},VisGenome \cite{krishna2017visual} & 2.1M & 37.4 \\
CoMVT \cite{seo2021look}                        & HowTo100M \cite{miech2019howto100m}  & 136M           & 39.5 \\
HD-VILA \cite{xue2022advancing}                 & HD-VILA-100M \cite{xue2022advancing} & 100M           & 40.0 \\
PMT \cite{peng2023efficient}                    & -                                    & -              & 40.3 \\
VQA\_T \cite{yang2021just}                      & -                                    & -              & 39.6 \\
VQA\_T \cite{yang2021just}                      & HowToVQA69M \cite{yang2021just}      & 69M            & 41.5 \\
ALPRO \cite{li2022align}                        & HowTo,WebVid                 & 5.5M                  & 42.1 \\
\midrule
\bf{VideoDistill\dag}                           & -                                    & -              & 42.7 \\
\bf{VideoDistill}                               & WebVid,HD-VILA,EgoCLIP               & 11M            & \bf{44.2} \\
\bottomrule

\end{tabular}%
}
\label{tab:MSRVTT-QA}
\vspace{12pt}
\centering
\tabcolsep=8pt
\caption{Comparison with SOTA methods on MSVD-QA.}
\resizebox{1\linewidth}{!}{%
\begin{tabular}{l c c | c}
\toprule
\bf {Method}         				            &\bf {Pretraining set}	        & \bf {Pairs}           & \bf {Acc}  \\
\midrule
HME \cite{fan2019heterogeneous}                 & -                             & - 		            & 33.7  \\
SSML \cite{amrani2021noise}                     & HowTo100M                     & 136M                  & 35.1 \\
HCRN \cite{le2020hierarchical}                  & -                             & - 		            & 36.1  \\
PMT \cite{peng2023efficient}                    & -                             & -                     & 41.8 \\
CoMVT \cite{seo2021look}                        & HowTo100M                    & 136M                  & 42.6 \\
SiaSamRea \cite{yu2021learning}                 & COCO,VisGenome               & 2.1M                  & 45.5 \\
ALPRO \cite{li2022align}                        & HowTo,WebVid                 & 5.5M                  & 45.9 \\
VQA\_T \cite{yang2021just}                      & HowToVQA69M                  & 69M                   & 46.3 \\
\midrule
\bf{VideoDistill\dag}                           & -                            & -                     & 46.2 \\
\bf{VideoDistill}                               & WebVid,HD-VILA,EgoCLIP       & 11M                  & \bf{49.2} \\
\bottomrule
\end{tabular}%
}
\label{tab:MSVD-QA}  
\end{table}

\section{Training Details}
\subsection{Pretraining Details}
\textbf{Pretraining Datasets.b} Our pretraining set consists of three parts: (1) 3M video-caption pairs randomly sampled from generic dataset WebVid10M \cite{bain2021frozen}. (2) 4.2M video-caption pairs randomly sampled from YouTube video dataset HD-VILA \cite{xue2022advancing}. We ensure the lengths of video clips sampled from WebVid10M and HD-VILA range from 10s to 30s. (3) 3.8M video-caption pairs from the 1st-person view dataset EgoCLIP \cite{lin2022egocentric}. Generally speaking, the 1st-person videos have more significant changes in perspective and orientation as the user moves around than the 3rd-person videos. Thus, they are helpful in releasing the potential of solving multiple events and multi-scale reasoning for VideoDistill.

\textbf{Implementation Details}
We resize all video clips (as well as downstream videos) to 256p while preserving the aspect ratio, then extract frames with 7.5 fps. We randomly sample 100 frames as input during pretraining and evenly sample 100 frames for downstream tasks. Finally, we augment input frames by random crop a $224\times 224$ region to increase input diversity.

In the video branch, we adopt CLIP-ViTB/16 \cite{radford2021learning} as the frame encoder. FS-Blocks and VB-Blocks have $L=3$ layers, a hidden size of $D=1024$. The number of attention heads equals 8 for all LA-Gates, self-attention layers, and spatial-temporal layers. We borrow spatial-temporal layers from FrozenInTime \cite{bain2021frozen}. We add a learnable temporal embedding for the input of the first FS-Block, a learnable temporal embedding, and a spatial embedding for the input of the first vision refinement block. We sparsely sample $K=16$ frames from 100 densely sampled frames as the input of vision refinement blocks for most experiments unless otherwise specified. In the text branch, we utilize the text encoder from CLIP with a maximum sequence length of 77.

For all experiments, we use AdamW optimizer with a learning rate of $3\times 10^{-5}$ and a weight decay of $1\times 10^{-3}$. Also, we employ a linear decay learning rate schedule with a warm-up strategy. We pretrain VideoDistill on 8 A100 GPUs with a batch size of 256 for 2 epochs (53 hours) to get our model applied to downstream tasks. Note that downstream performances may be further improved if we train the model for more epochs or customize better hyperparameters of the model architecture.

\begin{table*}[t]
\centering
\tabcolsep=18pt
\caption{Results on EgoMCQ multiple-choice test.}
\resizebox{1\linewidth}{!}{%
\renewcommand{\arraystretch}{0.8}
\begin{tabular}{l c c| c c}
\toprule
\multirow{2}{*}{\bf{Methods}}               &\multirow{2}{*}{\bf {Pretraining set}}	   & \bf {Pairs}       & \bf {Intra-video}        & \bf {Inter-video}  \\
                                            &                                            &             & ACC(\%)                  & ACC(\%)    \\
\midrule                                     
TimeSFormer+Distillbert                     & EgoCLIP                                    & 3.8M      & 85.5                     & 47.0\\
FrozenInTime \cite{bain2021frozen}          & EgoCLIP                                    & 3.8M      & 89.4                     & 51.5\\
EgoNCE w/Pos \cite{lin2022egocentric}       & EgoCLIP                                    & 3.8M      & 89.7                     & 53.6\\
EgoNCE w/Pos\&Neg \cite{lin2022egocentric}  & EgoCLIP                                    & 3.8M      & 90.6                     & 57.2\\
EgoVLP-v2 \cite{Pramanick_2023_ICCV}        & EgoCLIP                                    & 3.8M      & 91.0                     & 60.9\\
\midrule
\bf{VideoDistill\dag}                        & -                                         & -         & 92.0                     & 59.0 \\
\bf{VideoDistill}                            & WebVid,HD-VILA,EgoCLIP                    & 11M       & \bf{92.7}                & \bf{61.3}\\
\bottomrule
\end{tabular}%
}
\label{tab:EgoMCQ}
\vspace{12pt}
\centering
\tiny
\tabcolsep=22pt
\caption{Results on MSRVTT-multiple-choice test.}
\resizebox{1\textwidth}{!}{%
\renewcommand{\arraystretch}{0.8}
\begin{tabular}{l c c | c}
\toprule
\bf {Method}         				            &\bf {Pretraining set}	        & \bf {Pairs}           & \bf {Acc}  \\
\midrule
CT-SAN \cite{yu2017end}                         & - 		                    & -                     & 66.4     \\
MLB \cite{kim2016hadamard}                      & -                             & - 		            & 76.1  \\
JSFusion \cite{yu2018joint}          		    & -                             & - 		            & 83.4  \\
ActBERT \cite{zhu2020actbert}                   & HowTo100M                     & - 		            & 85.7  \\
ClipBert \cite{lei2021less}                     & COCO,VisGenome                & 2.1M                  & 88.2 \\
VideoCLIP\cite{xu2021videoclip}                 & HowTo100M 		            & 136M                  & 92.1 \\
HD-VILA \cite{xue2022advancing}                 & HD-VILA-100M                  & 100M                  & 97.1 \\
\midrule
\bf{VideoDistill}                               & WebVid,HD-VILA,EgoCLIP        & 11M                   & \bf{97.8} \\
\bottomrule
\end{tabular}%
}
\label{tab:MSRVTT-multiple-choice}
\end{table*}

\begin{table}[ht]
\caption{Comparison of text-to-video retrieval on MSR-VTT, 1k-A split. \dag denotes our model finetuned in a contrastive manner.} 
\centering
\resizebox{\linewidth}{!}{
\begin{tabular}{c c c | c c c}
\toprule
\bf {Method}         				    &\bf {PT-set}	                                & \bf {PT-pairs}  & \bf {R@1}     & \bf {R@5}     & \bf {R@10}\\
\midrule
CE\cite{liu2019use}                     & - 		                                    & -               & 20.9          & 48.8          & 62.4\\
UniVL\cite{luo2020univl}                & HowTo100M                          		    & 136M            & 21.2          & 49.6          & 63.1\\
ClipBERT\cite{lei2021less}              & COCO,VisGenome                        	    & 5.6M            & 22.0          & 46.8          & 59.9\\
FrozenInTime\cite{bain2021frozen}       & CC3M,WV2M,COCO                                & 6.1M            & 32.5          & 61.5          & 71.2\\
VideoCLIP\cite{xu2021videoclip}         & HowTo100M 		                            & 136M            & 30.9          & 55.4          & 66.8\\
HD-VILA \cite{xue2022advancing}         & HD-VILA-100M                                  & 100M            & \bf {35.6}    & 65.3          &\bf {78.0} \\
\midrule
\bf{VideoDistill\dag}       	        & WebVid,HD-VILA,EgoCLIP                        & 11M            & 32.8           & 63.5          & \underline{74.0} \\
\bf{VideoDistill}       	            & WebVid,HD-VILA,EgoCLIP                        & 11M         & \underline{33.4}  &\bf {70.1}     & 72.9 \\

\bottomrule
\end{tabular}%
}
\label{tab:MSRVTT}

\vspace{12pt}

\caption{Comparison of text-to-video retrieval on DiDeMo. \dag denotes generating the results of retrieval by direct similarity comparison like previous works (otherwise, by VTM head) during fine-tuning.} 
\centering
\resizebox{\linewidth}{!}{
\begin{tabular}{c c c | c c c}
\hline
\bf {Method}         				    &\bf {PT-set}	                                & \bf {PT-pairs}  & \bf {R@1}     & \bf {R@5}     & \bf {R@10}\\
\hline
HERO\cite{li2020hero}                   & TV\cite{lei2018tvqa},HowTo                    & 7.6M            & 2.1           & -             & 11.4 \\ 
S2VT\cite{venugopalan2014translating}   & COCO                          		        & -               & 11.9          & 33.6          & -\\
FSE \cite{zhang2018cross}               &  Sports-1M\cite{karpathy2014large}            & 1M              & 13.9          & 36.0          & -\\
CE\cite{liu2019use}                     & - 		                                    & -               & 16.1          & 46.1          & -\\
ClipBERT\cite{lei2021less}              & COCO,VisGenome                        	    & 5.6M            & 20.4          & 48.0          & 60.8\\
HD-VILA \cite{xue2022advancing}         & HD-VILA-100M                                  & 100M            & \bf {28.8}  & \underline{57.4}&\bf {69.1} \\
\hline
\bf{VideoDistill\dag}       	        & WebVid,HD-VILA,EgoCLIP                        & 11M            &\underline{28.0}& 57.1          & \underline{66.4} \\
\bf{VideoDistill}       	            & WebVid,HD-VILA,EgoCLIP                        & 11M            & 27.2           &\bf {61.6}     & 63.1 \\
\hline
\end{tabular}%
}
\label{tab:DiDeMo}
\end{table}

\begin{table*}[h]
\centering
\captionof{table}{Language-only QA results on the EgoTaskQA \textit{normal} split. (Gaussian inputs)}
\label{tab:without_vision_Gaussian}
\resizebox{0.8\textwidth}{!}{
\begin{tabular}{cccccccc}
\toprule
 \multirow{2}{*}{Category} & \multicolumn{2}{c}{VisualBERT~\cite{devlin2018bert}} & \multicolumn{2}{c}{HCRN (w/o vision)} & \multicolumn{2}{c}{\bf{VideoDistill} (w/o vision)}\\
 \cmidrule(lr){2-3}\cmidrule(lr){4-5} \cmidrule(lr){6-7}
 & Acc. & Change & Acc.& Change & Acc.& Change\\
 \midrule
world &  36.28 & \dec{-8.7\%} & 35.22 & \dec{-20.4\%} & 32.06 & \dec{-32.2\%} \\
intent & 35.02 & \dec{-21.3\%} & 34.93 & \dec{-29.8\%} & 26.56 & \dec{-49.4\%}\\
multi-agent & 20.58 & \dec{-21.7\%} & 19.17 & \dec{-38.9\%} & 18.58 & \dec{-49.7\%}\\
\midrule
descriptive & 34.55 & \dec{-17.7\%} & 33.58 & \dec{-22.8\%} & 29.45 & \dec{-36.7\%}\\
predictive & 24.75 & \dec{-18.5\%} & 24.3 & \dec{-33.5\%} & 19.93 & \dec{-50.7\%}\\
counterfactual & 41.3 & \dec{-1.6\%} & 40.4 & \dec{-15.8\%} & 39.51 & \dec{-20.4\%}\\
explanatory & 31.78 & \dec{-15.1\%} & 30.57 & \dec{-24.7\%} & 26.84 & \dec{-36.9\%}\\
\midrule
action & 15.72 & \inc{+4.6\%}  & 15.64 & \dec{-1.7\%} & 15.93 & \dec{-2.6\%}\\
object & 7.43 & \dec{-68\%} & 6.33 & \dec{-86.0\%} & 2.68 & \dec{-95.1\%}\\
state & 45.03 & \dec{-23.9\%} & 42.51 & \dec{-37.7\%} & 33.33 & \dec{-53.9\%}\\
change & 69.87 & \inc{+2.3\%} & 68.77 & \inc{+2.1\%} & 63.67 & \dec{-10.9\%}\\
\midrule
all & 33.92 & \dec{-10.6\%} & 32.51 & \dec{-23.0\%} &29.45 & \dec{-33.9\%}\\
\bottomrule
\end{tabular}
}
\end{table*}

\begin{table*}[h]
\caption{Performances on the EgoTaskQA \textit{indirect} split.}
\label{tab:indirect_split}
\centering
\resizebox{1\linewidth}{!}{%
\begin{tabular}{ccccccccccc}
\toprule
 & \multirow{2}{*}{Category} & \multirow{2}{*}{BERT} & \multirow{2}{*}[1ex]{HCRN} & \multirow{2}{*}{VisualBERT} & \multirow{2}{*}{PSAC} & \multirow{2}{*}{HME} & \multirow{2}{*}{HGA} & \multirow{2}{*}{HCRN} & \multirow{2}{*}{ClipBERT} &\multirow{2}{*}{\bf{VideoDistill\dag}}\\
    & & & (w/o vision) & & & & & & & \\
\midrule
\multirow{3}{*}{\rotatebox[origin=c]{90}{Scope}} & world & 34.96 & 33.61 & 40.00 & 44.74 & 35.91 & 31.29 & 44.04 & 26.51 &\textbf{47.82}\\
& intent & 23.56 & 23.98 & 36.02 & 48.38 & 31.73 & 20.42 & 47.02 & 14.66 &\textbf{49.61}\\
& multi-agent & 19.70 & 19.25 & 26.02 & \textbf{35.37} & 25.07 & 17.74 & 30.11 & 20.09 & \underline{35.04}\\
\midrule
\multirow{4}{*}{\rotatebox[origin=c]{90}{Type}} & descriptive & 33.09 & 30.73 & 38.9 & 43.36 & 34.48 & 29.01 & 42.02 & 24.35 &\textbf{45.13}\\
& predictive & 15.58 & 13.68 & 31.37 & 29.11 & 27.79 & 15.16 & 46.32 & 10.32 &\textbf{52.83}\\
& counterfactual & 34.59 & 34.75 & 37.63 & 39.94 & 35.07 & 33.01 & 43.64 & 26.29 & \textbf{43.97}\\
& explanatory & 27.38 & 28.11 & 32.75 & 42.53 & 29.16 & 24.00 & 39.69 & 22.46 &\textbf{43.75}\\
\midrule
\multirow{4}{*}{\rotatebox[origin=c]{90}{Semantic}} & action & 26.91 & 28.18 & 27.49 & 30.06 & 25.12 & 26.15 & 29.61 & 25.25 &\textbf{30.34}\\
& object & 2.808 & 4.13 & 22.63 & 30.97 & 19.08 & 7.02 & 32.20 & 10.49 & \textbf{45.97}\\
& state & 21.96 & 21.24 & 32.02 & 43.29 & 31.60 & 17.67 & 41.81 & 15.29 &\textbf{49.77}\\
& change & 55.28 & 50.71 & 55.59 & \textbf{57.20} & 47.65 & 47.22 & 56.27 & 35.26 &\underline{53.98}\\
\midrule
& all & 31.78 & 30.76 & 37.01 & 42.25 & 33.06 & 28.36 & 41.56 & 24.08 & \textbf{44.77}\\
\midrule
\multicolumn{2}{c}{Performance Change} & 6.4\% & 5.4\% & 2.4\% & 4.9\% & 17.7\% & 22.9\% & 1.5\% & 39.6\% & \textbf{0.25\%}\\
\bottomrule
\end{tabular}
}
\end{table*}

\subsection{Finetuning Details}
\textbf{Finetuning Datasets.} 

\textbf{EgoMCQ} \cite{lin2022egocentric} is a 1st-person Multiple-Choice Questions answering task. Each text query has five video candidates. It provides two criteria named Inter-video and intra-video accuracy. The former ensures the five video candidates come from different videos, and the latter collects candidates from the same video. The evaluation metric is accuracy.

\textbf{MSRVTT-QA} \cite{xu2017video} and \textbf{MSRVTT-multiple-choice test} \cite{yu2018joint} are two video question answering tasks basd on MSRVTT \cite{xu2016msr}. The former is open-ended, and the latter is multiple-choice. The evaluation metric is accuracy.

\textbf{MSVD-QA} \cite{xu2017video} is an open-ended question answering task with 1.9k short generic video clips. The evaluation metric is accuracy.

\textbf{EgoTaskQA} \cite{jia2022egotaskqa} is a long-form open-ended dataset with an average video length of 25s. It provides 15 categories of questions to evaluate models in detail. It also provides a version of the dataset (\textit{indirect} split) to reduce the usage of language shortcuts. The evaluation metric is accuracy.

\textbf{AGQA} \cite{grunde2021agqa} a long-form open-ended dataset contains 8 types of compositional spatiotemporal reasoning. The average video length is 30s. We use its v2 version, which has more balanced distributions, as the dataset creator recommended. The evaluation metric is accuracy.

\textbf{MSRVTT} \cite{xu2016msr} is 3rd-person video-text retrieval task. It contains 10K YouTube videos. We follow previous works \cite{yu2018joint,xue2022advancing}, finetuning SpaceCLIP on 9K videos and reporting results on the 1K-A test set. The evaluation metric is \textbf{R}@1, \textbf{R}@5, \textbf{R}@10.

\textbf{DiDeMo} \cite{anne2017localizing} consists of 10K Flickr videos and 40K manually annotated sentences. We use a standard split to fine-tune VideoDistill on the training set and report the result on the test set. The evaluation metric is \textbf{R}@1, \textbf{R}@5, \textbf{R}@10.

\textbf{Implementation Details.} For open-ended datasets MSRVTT-QA and MSVD-QA, EgoTaskQA, and AGQA, we take questions as the language input, then encode the answers in a one-hot fashion and train a two-layer MLP classification head over all answer candidates with a cross-entropy loss on the top of visual representation $v_{\rm{cls}}^{\rm{*}}$. For the multiple-choice dataset EgoMCQ, we respectively combine the five candidate videos with the question to form five input pairs, then choose the video corresponding with the maximum logit over the VTM head as the answer. For the multiple-choice dataset MSRVTT-multiple-choice test, we concatenate five answers with the question into five sentences, then choose the answer with the maximum logit over the VTM head. For text-to-video retrieval MSRVTT and DiDeMo, we provide two ways to realize retrieval. The first method is finetuning the module in a contrastive manner and choosing the answer with the highest similarity of $v_{\rm{cls}}^{\rm{*}}$ and $t_{\rm{cls}}$. The second method is choosing the answer with the highest VTM logits. We set the batch size to 128 and finetune the pretrained VideoDistill on 4 A100 GPUs. 

\section{Generic VideoQA}
We evaluate VideoDistill on the four commonly used VideoQA datasets: \textbf{MSRVTT-QA} \cite{xu2017video}, \textbf{MSVD-QA} \cite{xu2017video}, \textbf{EgoMCQ} \cite{lin2022egocentric} and \textbf{MSRVTT-multiple-choice test} \cite{yu2018joint}. 
\noindent
\textbf{Results.} In Table \ref{tab:MSRVTT-QA},\ref{tab:MSVD-QA},\ref{tab:EgoMCQ},\ref{tab:MSRVTT-multiple-choice}, the result of VideoDistill shows that our model outperforms existing methods on four tasks. On open-ended datasets MSRVTT-QA and MSVD-QA, we achieve 2.1\% and 2.9\% improvement over SOTA methods. Especially our from-scratch model outperforms previous large-scale pretrained models with 0.6\% gains. For multiple-choice datasets EgoMCQ and MSRVTT-multiple-choice test, the task setting is more like the retrieval and is more suitable for contrastive frameworks like HD-VILA\cite{xue2022advancing} and VideoCLIP\cite{xu2021videoclip}. Our model is still better than the SOTA methods. We find that VideoDistill achieves an improvement of 2.1\% on EgoMCQ Intra-video test, which is challenging since it ensures the five candidate answers are continuous clips with similar visual appearances. It shows that VideoDistill can better extract question-related visual semantics.

\section{Video-Text Retrieval}
Although VideoDistill is specially designed for VideoQA, we still evaluate it on text-to-video retrieval datasets MSRVTT \cite{xu2016msr} and DiDeMo \cite{anne2017localizing} to show its generalization power in Table \ref{tab:MSRVTT} and Table \ref{tab:DiDeMo}. 

\section{More quantitative results and ablations}
\textbf{The impact of LA-Gate.} To further demonstrate that LA-Gate can reduce the use of language prior, we eport the performance degradations of replacing visual inputs with Gaussian noise in Table \ref{tab:without_vision_Gaussian}. Similar to section 4.4 Table 3, we find that VideoDistill relies more on visual reasoning during the answer generation. 

 We also test VideoDistill on EgoTaskQA indirect split, which is motivated by the fact \cite{jia2022egotaskqa} that during task execution, actions, objects, and their changes are often strongly correlated. It leaves the chance for the model to perform well by simply over-fitting these strong correlations (language bias) without thorough task understanding. The indirect references can avoid these correlations. Table \ref{tab:indirect_split} shows that our VideoDistill has the least absolute performance change. It indicates that VideoDistill barely utilizes language bias in questions.

\textbf{The choice of the number of densely sampled frames.} We conduct the experiments in Table \ref{tab:different N} with $L=3$ and 16 encoded frames. We find that longer video clips (EgotaskQA) require a larger N to ensure we are not omitting the necessary information.
Nevertheless, too large N will damage the performance. One possible reason is a larger N needs more stacked frame sampling blocks. However, larger $L$ consumes more computing resources.

\textbf{Reasonable number of stacked layers $L$.} In Table \ref{tab:different L}, we set $N=100$ and simultaneously change $L$ for differentiable sparse sampling and vision refinement. We find too many layers still damage the performance since bigger $L$ dramatically improve the models' ability of fitting. Models will easily trapped in local minimums.

\textbf{The effectiveness of pretraining losses.} The designing concepts of pretraining losses are: MLM improves context reasoning by predicting the masked token. VTM and CL align visual and textual embeddings. Most of the time, applying one of VTM and CL is enough. This paper utilizes an incomplete CL to stabilize the training. Ablations on pretraining loss are shown in Table \ref{tab:Pretrain}. 

\begin{table}[h]
\vspace{-0.2cm}
\begin{minipage}{0.45\linewidth}
\caption{Sensitivity to densely sampled frames.}
\label{tab:different N}
\centering
\resizebox{\linewidth}{!}{
\begin{tabular}{c c c}
\toprule
\bf {N}   &\bf {EgoTaskQA}	 & \bf {MSRVTT-QA}      \\
\midrule
50    & 40.86 & 42.13 \\
100  & \textbf{45.02} & \textbf{44.20} \\
150  & 44.80 & 42.15 \\
200  & 42.12 & 41.10 \\
\bottomrule
\end{tabular}%
}
\end{minipage}\quad
\begin{minipage}{0.45\linewidth}
\caption{Sensitivity to the number of stacked blocks.}
\label{tab:different L}
\centering
\resizebox{\linewidth}{!}{
\begin{tabular}{c c c}
\toprule
\bf {L}   &\bf {EgoTaskQA}	 & \bf {MSRVTT-QA}      \\
\midrule
1   & 35.50 & 24.85 \\
3  & \textbf{45.02} & \textbf{44.20} \\
5  & 43.60 & 44.1 \\
8  & 42.18 & 43.59 \\
\bottomrule
\end{tabular}%
}
\end{minipage}
\end{table}

\begin{table}[h]
\tiny
\centering
\resizebox{0.\linewidth}{!}{%
\begin{tabular}{l c c c c}
\toprule
\multicolumn{1}{c}{\multirow{2}{*}{}} & \multicolumn{3}{c}{\bf {PT Tasks}}  &  \multicolumn{1}{c}{\multirow{2}{*}{\bf {MSRVTT-QA}}}     \\
\cmidrule(lr){2-4}
\multicolumn{1}{c}{} & \multicolumn{1}{c}{MLM} & \multicolumn{1}{c}{VTM} & \multicolumn{1}{c}{CL} &\multicolumn{1}{c}{}\\
\midrule
(a) & \ding{55}  & \ding{55}  & \ding{55}  & 42.7 \\
(b) & \checkmark & \ding{55}  & \ding{55}  & 43.4\\
(c) & \ding{55}  & \checkmark & \ding{55}  & 42.9  \\
(d) & \ding{55}  & \ding{55}  & \checkmark & 39.9 \\
(e) & \checkmark & \checkmark & \ding{55}  & 44.0  \\
(f) & \checkmark & \checkmark & \checkmark & \textbf{44.2}  \\
\bottomrule
\end{tabular}%
}
\vspace{-0.35cm}
\caption{Analysis of the effectiveness of pretraining tasks.}
\label{tab:Pretrain}
\end{table}

\section{Qualitative Results}
\begin{figure*}[ht]
\centering
\includegraphics[width=1\linewidth]{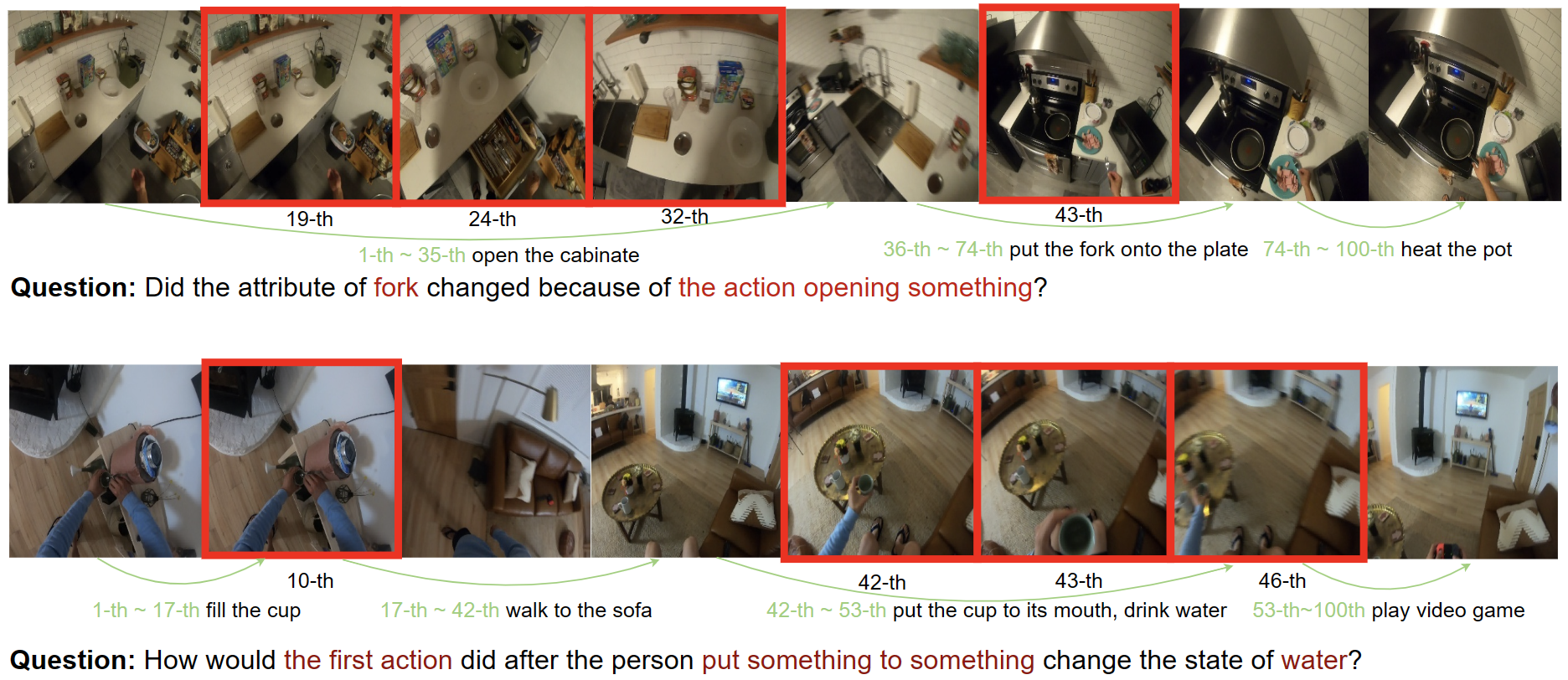}
\caption{Two instances from the four-frame variant}
\label{figure:4-frames}
\end{figure*}

\begin{figure*}[htbp]
\centering
\includegraphics[width=1\linewidth]{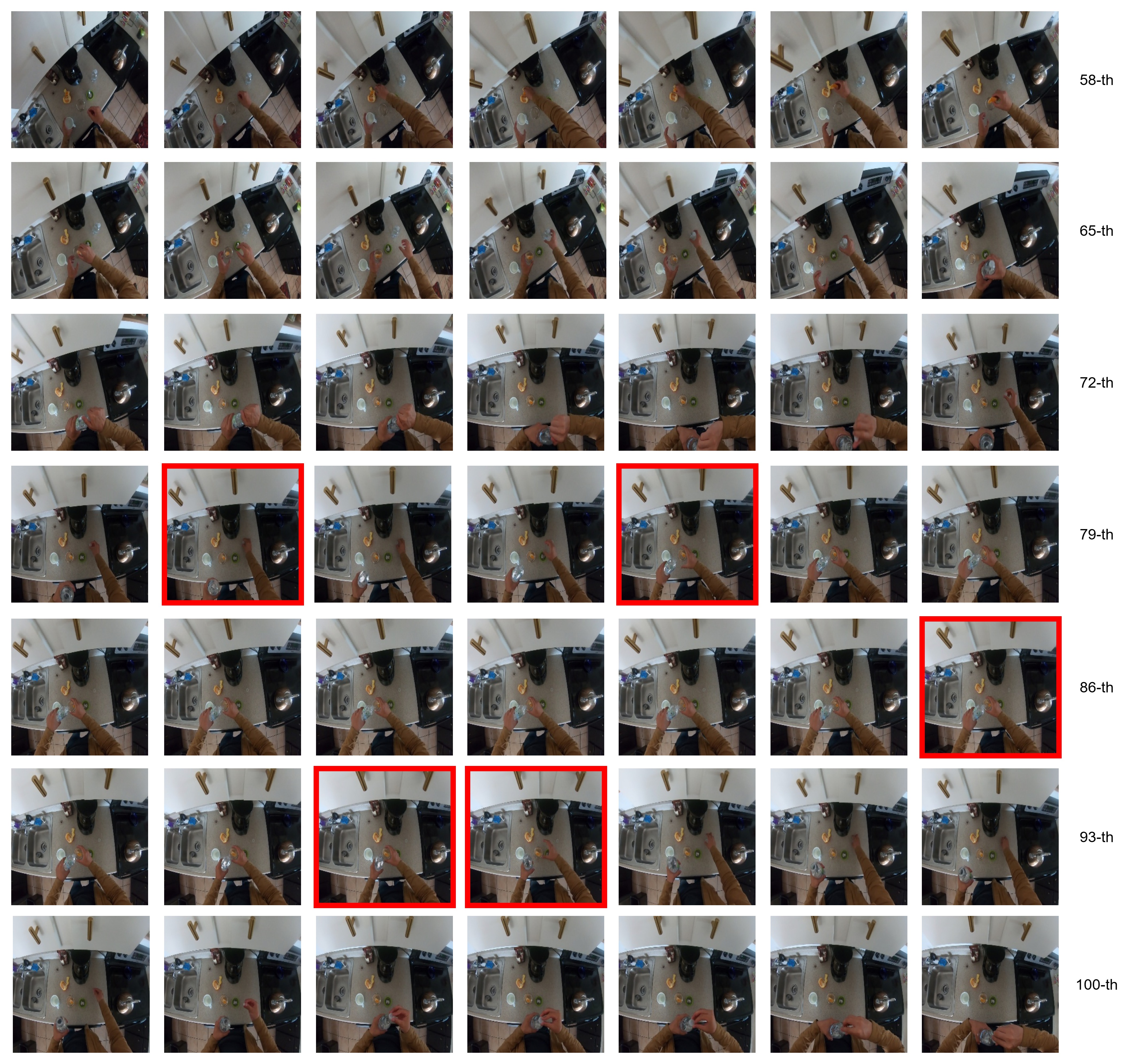}
\caption{A full instance from the 16-frame variant}

\label{figure:16-frames}
\end{figure*}

We visualize the result of our differentiable sparse sampling module. Specifically, we report two instances from a four-frame variant (the number of selected frames $K=4$) in Figure \ref{figure:4-frames} and a full instance from the sixteen-frame version used on downstream tasks in Figure \ref{figure:16-frames}. Note that models with $K>4$ allow duplicate selection, which means important frames can appear more than once in the $K$ selected frames.

\begin{figure*}[htbp]
\centering
\includegraphics[width=1\linewidth]{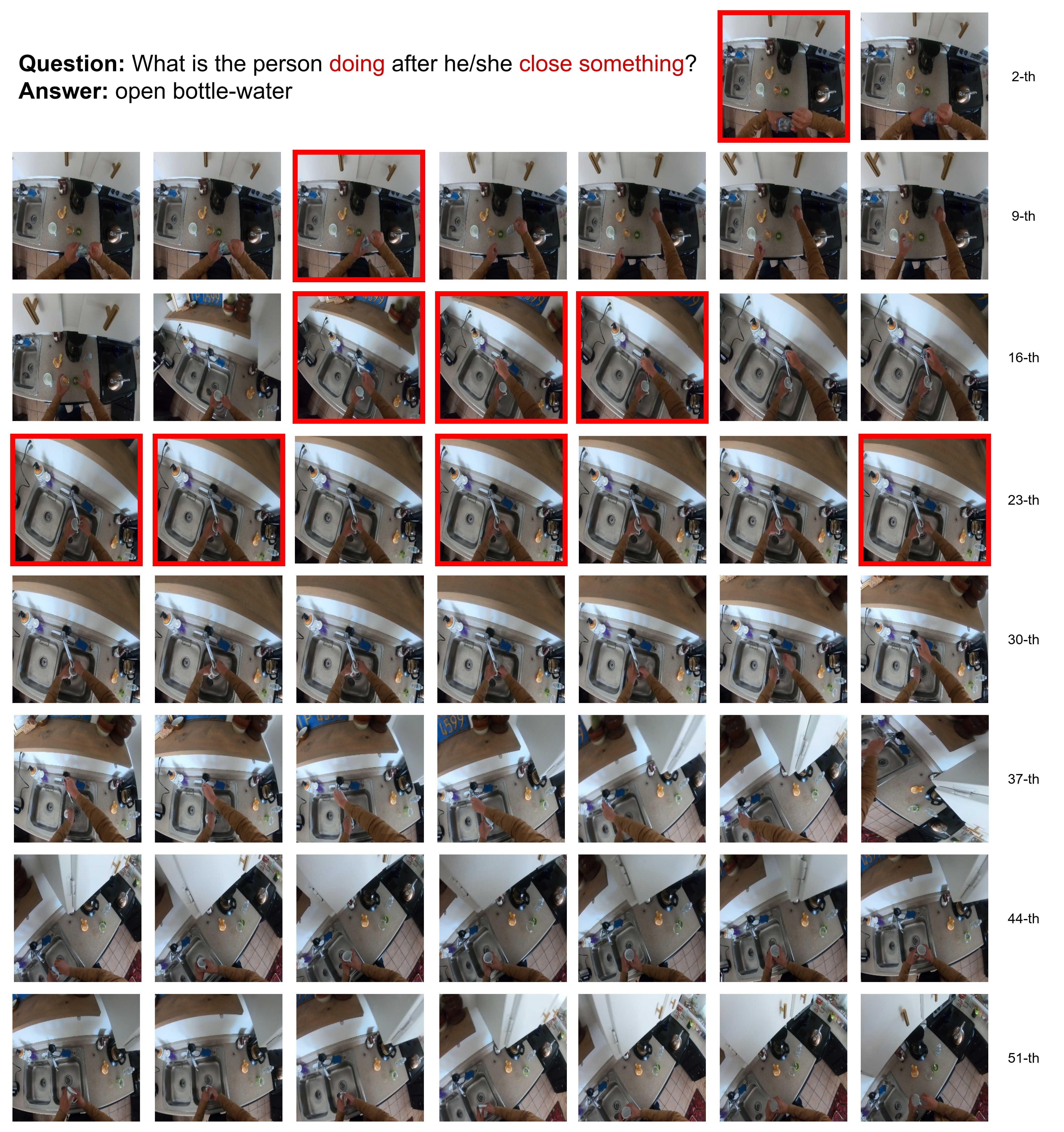}
\end{figure*}

